\documentclass [sigconf]{acmart}
\usepackage{multirow}
\usepackage{algorithmic}
\usepackage[ruled,vlined]{algorithm2e}
\usepackage{makecell} 
\usepackage[table]{xcolor}

\AtBeginDocument{%
  }

\setcopyright{acmlicensed}
\copyrightyear{2018}
\acmYear{2018}
\acmDOI{XXXXXXX.XXXXXXX}
\acmConference[Conference acronym 'XX]{Make sure to enter the correct
  conference title from your rights confirmation email}{June 03--05,
  2018}{Woodstock, NY}
\acmISBN{978-1-4503-XXXX-X/2018/06}

\begin{document}

\title{Log Anomaly Detection with Large Language Models via Knowledge-Enriched Fusion}






\author{Anfeng Peng}
\affiliation{%
  \institution{University of Pittsburgh}
  \city{Pittsburgh}
  \country{USA}
  }

\author{Ajesh Koyatan Chathoth}
\affiliation{%
  \institution{Eaton Corporation}
  \city{Pittsburgh}
  \country{USA}
  }

  
\author{Stephen Lee}
\affiliation{%
  \institution{University of Pittsburgh}
  \city{Pittsburgh}
  \country{USA}
  }
  

\renewcommand{\shortauthors}{Trovato et al.}
\begin{abstract}
System logs are a critical resource for monitoring and managing distributed systems, providing insights into failures and anomalous behavior. Traditional log analysis techniques, including template-based and sequence-driven approaches, often lose important semantic information or struggle with ambiguous log patterns. To address this, we present EnrichLog, a training-free, entry-based anomaly detection framework that enriches raw log entries with both corpus-specific and sample-specific knowledge. EnrichLog incorporates contextual information, including historical examples and reasoning derived from the corpus, to enable more accurate and interpretable anomaly detection. The framework leverages retrieval-augmented generation to integrate relevant contextual knowledge without requiring retraining.
We evaluate EnrichLog on four large-scale system log benchmark datasets and compare it against five baseline methods. Our results show that EnrichLog consistently improves anomaly detection performance, effectively handles ambiguous log entries, and maintains efficient inference. Furthermore, incorporating both corpus- and sample-specific knowledge enhances model confidence and detection accuracy, making EnrichLog well-suited for practical deployments.
\end{abstract}

\maketitle

\section{Introduction} 
System logs are a vital source of operational intelligence in large-scale distributed systems used for diagnosing system failures, identifying performance degradations, and potential security incidents.
In production environments, these logs are continuously streamed from thousands of microservices, containers, and infrastructure components, producing petabyte-scale datasets that capture rich temporal and contextual information about system behavior~\cite{AmazonLog,potluri2024deep,islam2025anomaly}. These logs contain critical system events that provide valuable evidence for detecting \textit{anomalies}, such as system malfunctions, misconfigurations, or security breaches~\cite{he2016experience, du2017deeplog,ahsan2018class,sujan2024enhancing}. As such, timely and accurate analysis of these logs is essential to prevent cascading failures and ensure system security and reliability. 

\begin{figure}
    \centering
    \includegraphics[width=3.2in]{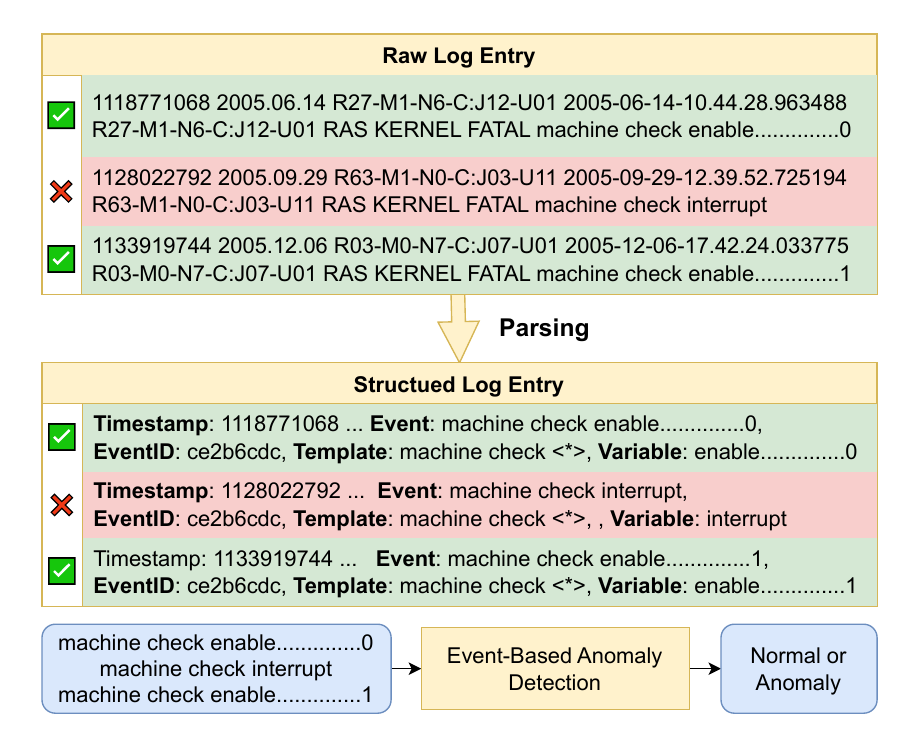}
    \caption{Example of ambiguous log templates in the BGL dataset, where both normal and anomalous log entries are mapped to the same template.}
    \label{fig:log_show}
\end{figure}

Despite their importance, anomaly detection from system logs remains challenging. Logs are often semi-structured, containing both structured metadata (e.g., timestamps) and unstructured free-text messages. In addition, log formats often differ across systems, services, and software versions, making it difficult to design a single model that generalizes effectively across diverse environments. To address this, logs are first processed using a log parser that extracts \textit{log templates} and converts raw entries into structured representations. These structured log templates are then analyzed using rule-based or machine learning-based techniques to detect anomalous patterns~\cite {du2017deeplog}. 
However, due to the wide variation in log formats across systems, separate models are often required for each environment, limiting their scalability.
Moreover, log parsers themselves can introduce errors. Syntax-based log parsers that use predefined or heuristics to extract log templates may map both normal and anomalous events to the same log template (see Figure~\ref{fig:log_show}). When downstream models analyze only on these templates, they risk missing contextual signals in the raw log text, which may cause false positives and missed anomalies. 


Recent advances in large language models (LLMs) have motivated researchers to explore template-driven log anomaly detection with retrieval-augmented generation (RAG)~\cite{he2024llmelog,Pan2023RAGLogLA, Zhang2024LeveragingRL,lewis2020retrieval}. In these approaches, annotated log templates are stored in a vector database, and when a new log entry arrives, relevant reference templates are retrieved and provided as context to the LLM for anomaly detection. This strategy improves performance compared to traditional rule-based or machine learning methods, as the LLM can reason over retrieved examples. However, because these methods still operate on log templates rather than raw log text, they inherit the same limitations of template-based pipelines. 
That is, a single template may correspond to both normal and anomalous events. 

To address these limitations, we present EnrichLog, a framework that directly leverages raw log text for anomaly detection while incorporating both corpus and sample-specific contextual knowledge. Corpus-specific knowledge consists of information available in the dataset documentation, including but not limited to definitions of anomalies, operational context requirements, and labeling methods. This knowledge provides a broad context for interpreting log events and guides the model in reasoning about what constitutes anomalous behavior. In contrast, sample-specific knowledge captures instance-level information for individual log entries, such as examples of anomalous or normal events along with explanations for their classification. In cases where a single log template corresponds to both normal and anomalous events, these distinctions are explicitly captured in the sample-specific knowledge. This allows the model to resolve ambiguities that arise from log templates. 

However, directly incorporating corpus-specific knowledge is non-trivial, as these documents are often unstructured and span multiple pages with extensive examples. Encoding and providing this information to LLMs is challenging because models struggle with long contexts, and even when all text is included, it is often unclear which portions are most relevant for a given log entry. Similarly, sample-specific knowledge requires careful structuring and enrichment to ensure that instance-level signals are effectively leveraged. EnrichLog addresses these challenges through a two-level knowledge fusion framework. First, corpus-specific knowledge is condensed into a global summary representation that captures dataset-level information relevant for anomaly detection. This global summary provides the model with essential high-level context without exceeding LLM context limits. Second, sample-specific knowledge is enriched with corpus-specific context using an LLM, emphasizing the distinctions present in individual log entries. This design makes EnrichLog particularly effective because it can (i) handle long-range context in a computationally efficient manner, (ii) focus on the most essential information from corpus- and sample-specific knowledge, and (iii) resolve ambiguities using rich signals from raw logs and contextual knowledge. Our contributions are summarized as follows:

\begin{itemize}
\item We present EnrichLog, a training-free, entry-based anomaly detection framework that enriches raw log entries with both corpus- and sample-specific contextual knowledge. Our sample-specific knowledge integrates historical examples and concise reasoning to enhance detection accuracy.
    \item We introduce a novel two-step inference strategy that balances efficiency and accuracy. The first step filters out confidently normal logs using a lightweight prompt, while the second step applies RAG for anomalous entries. This design reduces inference latency without sacrificing detection performance.
    \item We systematically evaluate EnrichLog across four large-scale system log datasets and against five baseline methods. Our results demonstrate that EnrichLog consistently improves anomaly detection performance, effectively handles ambiguous log entries, and maintains efficient inference. 
    \item Our study also provides a quantitative analysis of the impact of different types of contextual knowledge, embedding models, and log parsers. We show that concise, sample-specific enrichment enhances both confidence and accuracy. 


\end{itemize}

\section{Background}
\subsection{Log Parsing and Templates}
Logs are continuously generated and stored within a system, where each line corresponds to a log entry. A log entry typically comprises multiple attributes, such as timestamp, device identifier, and event information. Each log event is produced by administrator-defined code and adheres to a predefined template that is instantiated with specific parameters. These events capture the operational state of the system and may indicate deviations from normal behavior, referred to as anomalies.

To facilitate analysis, raw logs are preprocessed through log parsing 
\cite{Du2016SpellSP,He2017DrainAO,Zhu2018ToolsAB,Zhu2020LoghubAL,Jiang2023ALE}, which processes unstructured entries into standardized formats, shown in Figure~\ref{fig:log_show}. This standardization is important for capturing recurring patterns and providing a consistent basis for anomaly detection. Several approaches have been proposed for log parsing, ranging from syntax-based parsers methods~\cite{Du2016SpellSP,He2017DrainAO, hamooni2016logmine} to semantic-based parsers~\cite{huo2023semparser, le2023log, li2023did}. In this work, we primarily adopt Drain\cite{He2017DrainAO}, a popular syntax-based parser, but we also evaluate the efficacy of alternative parsers to extract log templates and better understand their impact on downstream anomaly detection performance.

Typically, log parsing extracts events from raw entries, clusters semantically similar events, and assigns them to corresponding templates. Consequently, each log entry can be abstracted into a template identifier, enabling the construction of template traces for log anomaly detection. However, this clustering process is not without challenges. In some cases, a single template may correspond to multiple labels, which introduces noise and ambiguity, ultimately degrading the performance of downstream anomaly detection tasks.

\subsection{Log-based Anomaly Detection}
Log-based anomaly detection can be broadly performed in two ways: (i) {session-based detection} and (ii) {entry-based detection}. These approaches differ in how they represent log data and formulate the anomaly detection task.

\textit{Session-based} techniques group log entries into sessions according to process identifiers (PIDs), time windows, or execution traces to identify anomalies. 
Early methods \cite{Du2017DeepLogAD,Lu2018DetectingAI,Brown2018RecurrentNN,Ede2022DEEPCASESC} relied on log parsing, where raw entries are mapped into template identifiers and then modeled as sessions. 
More recent language-model-based approaches \cite{he2024llmelog} extend this idea by incorporating semantic information alongside template ID sessions. Beyond template-based representations, some methods directly leverage raw log contents by embedding them into a continuous vector space \cite{Almodovar2024LogFiTLA}. Session-based methods capture contextual dependencies across logs, but they often lack fine-grained interpretability at the log-entry level.





\textit{Entry-based} approaches, in contrast, analyze individual log entries and formulate anomaly detection as a binary classification problem. 
These methods offer fine-grained detection and better interpretability, as each decision is tied directly to a specific log event. Some approaches can be considered template-based when the log parser preprocesses the logs and replaces raw entries with their corresponding templates. Recent work shows that enriching single-entry classification with contextual or semantic information can substantially improve accuracy \cite{Liu2024LogPromptPE,Zhang2024LeveragingRL}. However, entry-based approaches face scalability challenges, since every log entry is analyzed individually. 
In this paper, we focus on advancing the entry-based paradigm. We propose a knowledge-enriched framework that improves detection accuracy and interpretability while mitigating the computational overhead typically associated with LLM-based inference.



\subsection{Related Work}
Log anomaly detection has been extensively studied~\cite{chen2004failure, mariani2008automated, nagaraj2012structured, oprea2015detection, li2017flap, vinayakumar2017long}, with research evolving from traditional template-based methods~\cite{Du2017DeepLogAD} to deep learning–based approaches~\cite{zhang2016automated, zhang2019robust, guo2021logbert, liu2019log2vec}. Traditional methods relied heavily on log parsing and template identifiers. For example, DeepLog \cite{Du2017DeepLogAD} pioneered session-based detection using template ID sessions, while subsequent work explored CNN- and RNN-based architectures \cite{Lu2018DetectingAI,Brown2018RecurrentNN}. DeepCASE \cite{Ede2022DEEPCASESC} extended this line of research by incorporating human feedback into the detection pipeline. A major limitation of template-based methods, however, is the loss of semantic information present in raw logs. In contrast, our approach leverages the raw log content for anomaly detection to preserve semantic details to improve accuracy.

Subsequent approaches shifted toward richer semantic representations~\cite{Meng2019LogAnomalyUD, Lee2021LAnoBERTS,le2023log, yu2024loggenius, guan2024logllm, liu2024interpretable}. For example, LogAnomaly \cite{Meng2019LogAnomalyUD} proposed template2vec, embedding templates into word-based vectors to preserve semantic context. Similarly, LAnoBERTS \cite{Lee2021LAnoBERTS} introduced a BERT-based masked language model that leverages word-level embeddings to capture semantic dependencies within log sessions to enable anomaly detection. 

Language model–based anomaly detection can be broadly categorized into \textit{trainable} and \textit{training-free} approaches. Trainable methods construct task-specific models and train them under various paradigms, including unsupervised \cite{Lee2021LAnoBERTS}, weakly supervised \cite{He2025WeaklySupervisedLA}, semi-supervised \cite{Yang2021SemiSupervisedLA}, self-supervised \cite{Almodovar2024LogFiTLA}, and supervised \cite{Qi2023LogGPTEC, he2024llmelog}. These approaches consistently achieve state-of-the-art performance. However, their reliance on retraining makes them less adaptable to dynamic environments, where log distributions and messages can change over time. In contrast, EnrichLog adopts a training-free approach, eliminating the need for retraining while still leveraging contextual and historical knowledge to enhance anomaly detection.
 
Recent advances in large language models have made training-free approaches feasible. These methods leverage pretrained LLMs without any additional fine-tuning. Instead, task-specific semantics are incorporated through prompt engineering or retrieval-augmented generation (RAG), offering greater flexibility by eliminating the need for continuous retraining. Prompt-based methods, such as LogPrompt \cite{Liu2024LogPromptPE}, enrich prompts with reference logs to guide anomaly detection. 
However, there is a trade-off --- improving accuracy often necessitates longer prompts with additional reference logs, which increases inference latency and computational demand.

To address this limitation, RAG–based approaches (e.g., RAGLog \cite{Pan2023RAGLogLA}, LogRAG \cite{Zhang2024LeveragingRL}) dynamically retrieve the most relevant external knowledge during inference. For example, RAGLog stores reference logs in a knowledge base, while LogRAG augments prompts with retrieved templates. In contrast to these prior works, our approach systematically examines different types of contextual knowledge and quantifies their impact on anomaly detection performance. Moreover, rather than merely retrieving external information, our method enriches each log entry with sample-specific reasoning and historical examples to improve anomaly detection performance.

\section{EnrichLog Design}
\begin{figure*}
    \centering
    \includegraphics[width=7.1in]{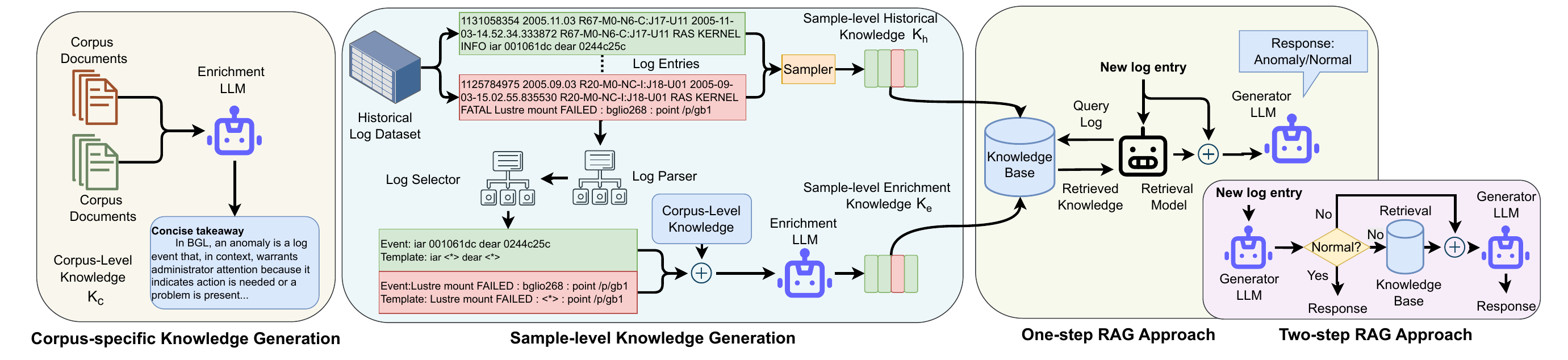}
    \caption{Overall architecture of the EnrichLog framework. }
    \label{fig:Architecture_overall}
\end{figure*}

In this section, we describe the details of our approach. The key goal of our approach in this work is to explore how contextual knowledge can be leveraged for log anomaly detection without requiring model training. To this end, we systematically explore optimization opportunities to improve both the effectiveness and efficiency of large language model (LLM)-based detection. The overall architecture of our proposed framework, EnrichLog, is illustrated in Fig.~\ref{fig:Architecture_overall}. 
The framework consists of two key components: (a) Contextual knowledge generation, (b) RAG-based anomaly detection process. Contextual knowledge generation constructs both corpus-specific and sample-specific knowledge to provide domain and instance-specific semantics for log entries. This sample-specific knowledge is encoded and stored in a vector database, also called a knowledge base, to facilitate efficient retrieval. In the retrieval-augmented process, the system retrieves relevant contextual knowledge and integrates it with the raw log input to perform anomaly detection. We discuss each component in detail below.

\subsection{Contextual Knowledge Generation}
LLMs are trained on large amounts of publicly available data, which encodes the knowledge acquired during pretraining. While this enables strong generalization across many tasks, LLMs often lack the up-to-date or domain-specific knowledge required for accurate anomaly detection in system logs. Logs are highly domain-dependent and capture environment-specific behaviors that are rarely represented in pretraining corpora. As a result, relying solely on a pretrained model to analyze log entries can lead to poor performance. A common solution is to integrate an external knowledge base that encodes domain and corpus-specific semantics, thereby enriching the model’s reasoning with context unavailable in its parameters. However, such knowledge bases are often substantially large and span multiple pages. This can easily exceed the LLM's context window, making integrating this knowledge challenging. To address this, we first generate \textit{corpus-specific knowledge}, consisting of a concise summary of the relevant corpora. This corpus-specific knowledge is then used to generate \textit{sample-specific knowledge}, which informs the anomaly detection process. 

\subsubsection{Corpus-specific Knowledge}
    

System logs exhibit recurring structures and patterns, including standardized error codes that are often documented in external knowledge sources. 
To achieve this, we provide the LLM with a structured prompt that instructs it to produce a concise summary capturing the essential semantic information. This process is illustrated in Figure~\ref{fig:Architecture_overall}. Each summarization task may include one or more documents, enabling the generation of a corpus-specific knowledge representation for the dataset. For each dataset, we provide the LLM with the relevant log documentation, including anomaly definitions and identification methods. The LLM then generates a concise knowledge summary $K_c$ of the documents. 
\begin{equation}
    K_c =  \Theta(C)
\end{equation}
where $\Theta(.)$ denotes the LLM, $C$ represent the corpus documents, $c$ indicates the corpus, and $M$ is the number of property.  
This design also reduces the number of tokens passed to the LLM, thereby improving computational efficiency.


\subsubsection{Sample-specific Knowledge}
While corpus-specific knowledge captures global patterns and templates, sample-specific knowledge focuses on the local context of individual log entries. The objective is to provide a concise explanation of why a particular log entry is normal or anomalous, highlighting the key reason behind its classification. For instance, the explanation may indicate whether the system is operating as expected or exhibiting abnormal behavior. In cases where a log template is ambiguous, multiple explanations are provided to account for possible interpretations to ensure that the enrichment reflects all plausible reasoning paths. Our sample-specific knowledge consists of two components: (i) sample enrichment and (ii) historical samples, both of which are used for anomaly detection. Sample enrichment provides detailed reasoning about the log entry classification. Historical samples, on the other hand, consist of actual past log entries and serve as concrete examples.

The sample-specific enrichment process is performed in three main steps (see Figure~\ref{fig:Architecture_overall}). First, a \textit{log parser} extracts the structural template of each log entry. This reduces the number of unique entries the anomaly detection model must process, improving efficiency and scalability. Representing logs as templates also makes it easier to identify recurring patterns and typical system behaviors, which helps in distinguishing normal entries from anomalies. Next, a \textit{log selector} identifies a representative log template for the log entries to enrich. Since a single log template may correspond to multiple similar log events, selecting only representative samples reduces the amount of information stored in the knowledge base. Finally, we prompt the LLM to generate concise, human-readable explanations that articulate why each selected sample is normal or anomalous to generate sample-specific enriched knowledge $K_e$ as follows. 
\begin{equation}
    K_e = \Theta (x, y, K_c)
\end{equation}
where $\Theta(.)$ denotes the LLM that generates $K_e$ from log entry $x$, the ground-truth label $y$ (denoting whether $x$ is anomalous or normal), and the corpus-specific knowledge $K_c$. The outputs are stored in a knowledge base, which is later used in the retrieval process to provide contextual guidance during anomaly detection.

The historical samples, $K_h$, consist of previously observed structured log entries (including log events and labels) that serve as concrete examples to guide anomaly detection. For each log template type, we subsample $x \in X$. 
\begin{equation}
K_h = \text{Sampler}(X)    
\end{equation}
where Sampler randomly selects and returns a subset of historical log entries for each log template type. The resulting $K_h$ is then stored in the knowledge base.

\subsection{RAG Anomaly Detection}
The RAG-based detection process in EnrichLog integrates raw log events, historical logs, and enriched knowledge to perform anomaly detection using a generative LLM. Given a new log entry $x$, the system first retrieves relevant sample-specific knowledge ($z_e\in K_e$ and $z_h\in K_h$) to guide the generation of the target anomaly label $y$.

Formally, EnrichLog uses a retriever model $R$ to retrieve the most relevant sample-specific enrichment knowledge $z_e = R(K_e | x)$ given a log entry $x$. 
Additionally, the retriever model retrieves the sample-specific historical knowledge $z_h = R(K_h | x)$ to provide concrete past examples to guide classification. 
The generator LLM 
$\Theta(y | x, z_e, z_h)$  produces the anomaly label in an autoregressive manner, generating each token sequentially conditioned on the input log $x$, the retrieved sample-specific knowledge $z_e$ and $z_h$, and previously generated tokens $y_{1:i-1}$. The token-specific probability of generating the label sequence $y = (y_1, \dots, y_{|y|})$ is:

\begin{equation}
p_{\text{EnrichLog}}(y|x, z_e, z_h) \approx \prod_{i=1}^{|y|} p_\theta(y_i | x, z_e, z_h, y_{1:i-1})
\end{equation}
Here, model $\Theta(.) $ is used to compute the probability $p_\theta$, and each $y_i$ represents the $i$-th token in the anomaly label sequence. 

This token-specific formulation also allows the computation of a confidence score for the prediction. Specifically, the confidence of a predicted label is computed as follows. Given the per-token logits from the LLM, we can directly use the probability of generating each candidate answer as its confidence. For a candidate label $\hat{y} \in A$, where $A$ is the set of possible class labels (e.g., ${\text{normal}, \text{anomaly}}$), the confidence is calculated as:
\begin{equation}
\text{Confidence}(\hat{y} \mid x, z_e, z_h) =
\frac{\exp\Big(\text{logits}_{\hat{y}} / \tau\Big)
}{\sum_{y_j \in A} \exp\Big(\text{logits}_{y_j} / \tau\Big)}   
\label{exp:confidence}
\end{equation}
where $\text{logits}_{{y}}=\text{logits}_\theta({y} \mid x, z_e, z_h)$ denotes the LLM's output logit for label ${y}$, and $\tau$ is a sampling temperature parameter of language models.

\begin{algorithm}[ht]
\small
\caption{EnrichLog: Contextual Knowledge Enrichment and RAG-based Anomaly Detection}
\label{alg:enrichlog}
\KwIn{Corpus documents $\mathcal{D}$, historical log entries $\mathcal{X}_h$, new log entries $\mathcal{X}_q$, LLM $\Theta$, retriever $R$, candidate labels $A$}
\KwOut{Predicted anomaly labels $\hat{y}$ and confidence scores}

\BlankLine
\textbf{Step 1: Generate corpus-specific knowledge $K_c$} \\
    $K_c \gets \Theta(D)$ 

\BlankLine
\textbf{Step 2: Generate sample-specific enrichment knowledge $K_e$} \\
$K_e \gets \emptyset$ \\
$\mathcal{X}_h',\mathcal{T} \gets \text{LogParser}(\mathcal{X}_h)$ \tcp*{$\mathcal{T}$ are log templates, $\mathcal{X}_h'$ are structured log entries}
$\mathcal{X}_h'$ = $\bigcup_{(l, t)} \mathcal{X}_h'^{(l,t)}$ \tcp*{Selector groups $\mathcal{X}_h'$ by label and template}

\ForEach{(log entry $x_e$, label $y) \in \mathcal{X}_h' $}
{
    $k_e \gets \Theta(x_e, K_c, y)$ \tcp*{Sample-specific explanation, $x_e$ is the event.}
    $K_e \gets K_e \cup \{k_e\}$
}
Store $K_e$ in the knowledge base
\BlankLine
\textbf{Step 3: Store sample-specific historic knowledge $K_h$} \\
$K_h \gets $ Sampler($\mathcal{X}_h$) \\
Store $K_h$ in the knowledge base
\BlankLine
\textbf{Step 4: RAG-based anomaly detection} \\
\ForEach{new log entry $x_q \in \mathcal{X}_q$}{
    \If{One-step mode}{
        $z_e \gets R(K_e \mid x_q)$ \tcp*{Retrieve sample-specific enrichment knowledge relevant to $x_q$}
        $z_h \gets R(K_h \mid x_q)$ \tcp*{Retrieve sample-specific historical knowledge relevant to $x_q$}
        $\hat{y} \gets \Theta(x_q, z_e, z_h)$ \tcp*{Generate anomaly label autoregressively with full context}
    }
    \ElseIf{Two-step mode}{
        $\hat{y}^{(1)} \gets \Theta(x_q)$ \tcp*{Initial classification using only raw log event}
        \If{$\hat{y}^{(1)} == \text{normal with high confidence}$}{
            $\hat{y} \gets \text{normal}$ \tcp*{No further process}
        } 
        \Else{
            $z_e \gets R(K_e \mid x_q)$ \tcp*{Retrieve sample-specific enrichment knowledge relevant to $x_q$}
            $z_h \gets R(K_h \mid x_q)$ \tcp*{Retrieve sample-specific historical knowledge relevant to $x_q$}
            $\hat{y} \gets \Theta(x_q, z_e, z_h)$  \tcp*{Refine classification using contextual knowledge}
        }
    }


    \tcp{Compute confidence from logits over candidate labels}
    \ForEach{candidate label $\hat{y}_i \in A$}{
        $\text{Conf}(\hat{y}_i \mid x_q, z_e, z_h) \gets \frac{\exp(\text{logits}_\theta(\hat{y}_i \mid x_q, z_e, z_h)/\tau)}{\sum_{y_j \in A} \exp(\text{logits}_\theta(y_j \mid x_q, z_e, z_h)/\tau)}$
    }
}
\Return{$\hat{y}$, $\text{Conf}(\hat{y})$}
\end{algorithm}

{\bf One-step Approach.}
Algorithm~\ref{alg:enrichlog} presents the EnrichLog framework for training-free anomaly detection using retrieval-augmented LLMs. In the one-step approach, for each query log entry, the retriever retrieves the top relevant sample-specific knowledge entries. The generator LLM then produces the anomaly label autoregressively, conditioned on the sample-specific knowledge entries.

{\bf Two-step Approach.}
While incorporating contextual knowledge improves anomaly detection precision, it also increases prompt length and computation time, particularly affecting the time-to-first-token. 
When contextual knowledge is not used, the model typically achieves high recall but low precision, causing some normal logs to be misclassified as anomalies. To address this, we propose a two-step approach. In the first step, the pretrained LLM classifies log entries using only the raw log text, efficiently filtering out entries that are confidently normal and reducing the number of logs requiring more expensive processing. In the second step, only log entries classified as anomalous or uncertain are processed with the full retrieval-augmented prompt, which incorporates both corpus-specific and sample-specific knowledge. 
This two-step approach accelerates inference because the first step is fast, relying solely on the basic prompt, while the slower second step is executed only for logs that require additional context for accurate classification.

\section{Experimental Methodology}

\subsection{Datasets}

We conduct experiments on four benchmark datasets for log anomaly detection: \textbf{BGL}, \textbf{Thunderbird}, \textbf{Spirit}, and \textbf{Liberty}~\cite{Oliner2007WhatSS}. 


\textbf{Blue Gene/L (BGL)} dataset is a system log collected from the Blue Gene/L supercomputer at Lawrence Livermore National Laboratory (LLNL)~\cite{Oliner2007WhatSS}. It consists of 4,747,963 log entries, each manually labeled as normal or anomalous. 

\textbf{Thunderbird} is a system log dataset collected from a large-scale cluster at Sandia National Laboratories (SNL)~\cite{Oliner2007WhatSS}, containing 211,212,192 raw log entries.

\textbf{Spirit} is a system log dataset collected from the Spirit supercomputer at Sandia National Laboratories (SNL)~\cite{Oliner2007WhatSS}. It consists of 272,298,969 raw log entries.

\textbf{Liberty} is a system log dataset collected from the Liberty supercomputer. It contains 265,569,231 log entries, including both normal and anomalous entries.

We split each dataset into training, validation, and test sets at a 6:2:2 ratio. The training set serves as the historical log dataset. We randomly sample 20,000 log entries from each training set to serve as the sample-level historical knowledge. The validation set is reserved for future comparison on the supervised approach. Following prior work~\cite{Zhu2020LoghubAL}, for evaluation, we sample 2,000 log entries from each test set, with 
anomaly rates 8\% for BGL, 1.4\% for Thunderbird, 67\% for Spirit, and 37\% for the Liberty dataset.




\subsection{Baseline Methods}
We compare EnrichLog against several representative baselines.

\textbf{Baseline}. We evaluate the effectiveness of different pre-trained LLMs for anomaly detection using their intrinsic knowledge, without any external enrichment. Specifically, we explore five open-source LLM models of varying sizes: small (Llama-3.2-1B-Instruct), medium (Gemma-3-4b-it and Qwen3-4B-Instruct-2507), and large (Mistral-7B-Instruct-v0.3 and Llama-3.1-8B-Instruct). We refer to them as Llama1B, Gemma3-4b, Qwen3-4B, Mistral-7B, and Llama8B, respectively, in the following sections.
They enable us to study the impact of model size on both anomaly detection performance and computational efficiency.

\textbf{Corpus-only}. This has the same pipeline as the \textbf{Baseline}, but corpus-specific knowledge is provided in the prompt. 

\textbf{LogPrompt \cite{Liu2024LogPromptPE}}. We use the in-context mode of this method. It augments the prompt with a fixed set of reference logs and performs binary classification of new log entries. Due to context window limitations and speed, we randomly select 100 logs from the historical knowledge pool to include in the prompt.

\textbf{RAGLog \cite{Pan2023RAGLogLA}.} RAGLog maintains a knowledge base of historical log entries and retrieves the top-K most relevant entries for each new query. Since the original code is not open-sourced, we re-implemented the retrieval–prompt integration for replication. The prompt includes the task, retrieved log events,  and the new log event.

\textbf{LogRAG \cite{Zhang2024LeveragingRL}}. This method stores log tokens and templates in a database, with anomaly detection results reproduced using the official open-source implementation.

We also explored other log anomaly detection frameworks, including MidLog~\cite{Yuan2025MidLogAA}, PLELog~\cite{yang2021plelog}, and LLMeLog~\cite{he2024llmelog}. However, these approaches rely on sequences of multiple log entries to detect anomalies and thus fail when only a single log entry is available. As such, we do not include their results in this work.

We also examined recent studies that leverage proprietary large language models (LLMs), such as OpenAI’s ChatGPT. But these models are continuously updated~\cite{openai2025modelrelease}. Meanwhile, the prompts used in RAGLog~\cite{Pan2023RAGLogLA} were not disclosed. To ensure fairness and reproducibility, we instead employ open-source LLMs with frozen weights and prompts the same as ours. 
We plan to release our model weights and code with the camera-ready version paper.



\subsection{Metrics}
Similar to prior work \cite{Yang2021SemiSupervisedLA}, we adopt three standard metrics: precision, recall, and F1-score.

\textbf{Precision} 
is the percentage of correctly detected anomalous log sequences among all sequences predicted as anomalous by the model, defined as $P = TP/ (TP+FP)$, where TP is the number of true positives and FP is the number of false positives.

\textbf{Recall} is the percentage of actual anomalous log sequences that are correctly identified by the model, defined as $R=(TP / (TP + FN)$. 



\textbf{F1-score} represents the harmonic mean of precision and recall.

In addition, we use \textbf{confidence score} as defined in (\ref{exp:confidence}) to quantify the certainty of model predictions. Since LLMs generate tokens probabilistically based on next-token distributions, these probabilities can be interpreted as a measure of confidence.

\section{Results}





\subsection{Baseline Performance Comparison}
\begin{table*}[th]
    \centering
    \footnotesize
    \renewcommand{\arraystretch}{1.2}
    \begin{tabular}{l l |ccc| ccc| ccc| ccc| ccc}
        \toprule
        Dataset & Method &\multicolumn{3}{c}{Llama-3.2-1B-Instruct}&\multicolumn{3}{c}{Gemma-3-4b-it}&\multicolumn{3}{c}{Qwen3-4B-Instruct}&\multicolumn{3}{c}{Mistral-7B-Instruct-v0.3}&\multicolumn{3}{c}{Llama-3.1-8B-Instruct} \\
        \cmidrule(lr){3-5} \cmidrule(lr){6-8} \cmidrule(lr){9-11} 
        \cmidrule(lr){12-14} \cmidrule(lr){15-17}
        & & P & R & F1 & P & R & F1 & P & R & F1 & P & R & F1 & P & R & F1 \\
        \midrule

        \multirow{7}{*}{BGL} 
        & Base          & 7.95 & 100.00 & 14.73 & 8.70 & 100.00 & 16.01 & 21.17 & 100.00 & 34.95 & 24.31 & 99.37 & 39.06 & 23.87 & 99.37 & 38.49 \\
        & LogPrompt   & 30.17&45.91&\underline{36.41}&41.98&77.36&54.42&29.03&90.57&43.97&10.15&100.00&18.42&68.98&81.13&74.57 \\
        & LogRAG        & 50.32 & 100.00 & \textbf{66.95} & 37.50 & 1.89 & 3.59 & 54.55 & 3.77 & 7.06 & 79.80 & 99.37 & 88.52 & 0.00 & 0.00 & 0.00 \\
        & RAGLog        & 0.00 & 0.00 & 0.00 & 74.53 & 99.37 & \underline{85.18} & 80.51 & 98.74 & 88.70 & 92.44 & 100.00 & 96.07 & 61.45 & 96.23 & \underline{75.00} \\
        & Corpus-only   & 9.65 & 6.92 & 8.06 & 16.71 & 39.62 & 23.51 & 21.97 & 99.37 & 35.99 & 22.09 & 82.39 & 34.84 & 22.53 & 86.16 & 35.72 \\
        \rowcolor{gray!20}
       \cellcolor{white} & \textbf{EnrichLog}     & 0.00 & 0.00 & 0.00 & 99.37 & 99.37 & \textbf{99.37} & 98.73 & 98.11 & \underline{98.42} & 98.75 & 99.37 & \underline{99.06} & 92.81 & 97.48 & \textbf{95.09} \\
        \rowcolor{gray!20}
       \cellcolor{white} & \textbf{EnrichLog*}    & 0.00 & 0.00 & 0.00 & 99.37 & 99.37 & \textbf{99.37} & 99.36 & 98.11 & \textbf{98.73} & 99.37 & 99.37 & \textbf{99.37} & 92.81 & 97.48 & \textbf{95.09} \\
        \midrule

        \multirow{6}{*}{Thunderbird} 
        & Base          & 1.40 & 100.00 & 2.77 & 3.09 & 100.00 & 6.00 & 2.08 & 100.00 & 4.07 & 3.93 & 100.00 & 7.56 & 3.46 & 100.00 & 6.68 \\
        & LogPrompt     &28.21&39.29&\textbf{32.84}&4.75&100.00&9.08&6.97&100.00&13.02&1.84&100.00&3.61&22.76&100.00&37.09 \\
        & LogRAG        & 13.93 & 100.00 & \underline{24.45} & 0.00 & 0.00 & 0.00 & 0.00 & 0.00 & 0.00 & 62.22 & 100.00 & \underline{76.71} & 0.00 & 0.00 & 0.00 \\
        & RAGLog        & 3.23 & 14.29 & 5.26 & 18.79 & 100.00 & \underline{31.64} & 20.59 & 100.00 & 34.15 & 16.47 & 100.00 & 28.28 & 41.18 & 100.00 & \textbf{58.33} \\
        & Corpus-only   & 1.76 & 100.00 & 3.45 & 3.95 & 100.00 & 7.60 & 2.59 & 100.00 & 5.05 & 3.84 & 100.00 & 7.39 & 3.74 & 100.00 & 7.21 \\
        \rowcolor{gray!20}
       \cellcolor{white} & \textbf{EnrichLog}     & 10.53 & 8.00 & 9.09 & 100.00 & 100.00 & \textbf{100.00} & 90.32 & 100.00 & \underline{94.92} & 93.33 & 100.00 & \textbf{96.55} & 
        20.74 & 100.00 & 34.36 \\
        \rowcolor{gray!20}
       \cellcolor{white} & \textbf{EnrichLog*}    & 25.00 & 3.57 & 6.25 & 100.00 & 100.00 & \textbf{100.00} & 100.00 & 100.00 & \textbf{100.00} & 93.33 & 100.00 & \textbf{96.55} & 
        23.93 & 100.00 & \underline{38.62} \\
        \midrule

        \multirow{6}{*}{Spirit} 
        & Base          & 66.75 & 100.00 & 80.06 & 79.06 & 100.00 & 88.31 & 71.86 & 100.00 & 83.63 & 82.90 & 100.00 & 90.65 & 82.12 & 99.92 & 90.15 \\
        & LogPrompt     & 97.19&98.72&\textbf{97.95}&99.62&98.80&99.21&97.14&99.55&\underline{98.33}&98.81&99.25&99.03&97.13&99.17&98.14 \\
        & LogRAG        & 96.16 & 99.62 & \underline{97.86} & 58.33 & 0.53 & 1.04 & 99.24 & 19.50 & 32.60 & 98.45 & 100.00 & 99.22 & 35.00 & 0.53 & 1.03 \\
        & RAGLog        & 58.63 & 13.50 & 21.95 & 88.87 & 40.74 & 55.86 & 91.14 & 40.89 & 56.45 & 94.51 & 99.47 & 96.93 & 99.24 & 98.57 & 98.91 \\
        & Corpus-only   & 66.65 & 100.00 & 79.99 & 80.67 & 39.46 & 53.00 & 93.48 & 100.00 & 96.63 & 84.26 & 100.00 & 91.46 & 85.99 & 99.47 & 92.24 \\
        \rowcolor{gray!20}
       \cellcolor{white}  & \textbf{EnrichLog}     & 100.00 & 2.85 & 5.54 & 99.78 & 100.00 & \underline{99.89} & 99.78 & 100.00 & \textbf{99.89} & 99.78 & 100.00 & \underline{99.89} & 98.81 & 100.00 & \textbf{99.40} \\
        \rowcolor{gray!20}
       \cellcolor{white} & \textbf{EnrichLog*}    & 97.78 & 3.30 & 6.39 & 99.93 & 100.00 & \textbf{99.96} & 99.78 & 100.00 & \textbf{99.89} & 99.93 & 100.00 & \textbf{99.96} & 98.81 & 99.47 & \underline{99.14} \\
        \midrule

        \multirow{6}{*}{Liberty} 
        & Base          & 37.95 & 100.00 & 55.02 & 47.38 & 100.00 & 64.29 & 39.93 & 100.00 & 57.07 & 48.81 & 100.00 & 65.60 & 46.77 & 100.00 & 63.73 \\
        & LogPrompt     &95.85&97.36&\underline{96.60}&97.68&100.00&\underline{98.83}&99.20&98.55&98.88&97.29&99.34&98.31&98.28&97.76&98.02 \\
        & LogRAG        & 88.98 & 100.00 & \textbf{94.17} & 15.56 & 0.92 & 1.74 & 98.92 & 96.18 & 97.53 & 97.94 & 100.00 & 98.96 & 6.25 & 0.13 & 0.26 \\
        & RAGLog        & 0.00 & 0.00 & 0.00 & 96.08 & 100.00 & 98.00 & 96.44 & 100.00 & 98.19 & 96.56 & 100.00 & 98.25 & 99.60 & 99.34 & \textbf{99.47} \\
        & Corpus-only   & 37.95 & 100.00 & 55.02 & 49.80 & 98.81 & 66.23 & 89.32 & 99.21 & 94.01 & 49.58 & 100.00 & 66.29 & 48.25 & 99.60 & 65.00 \\
        \rowcolor{gray!20}
       \cellcolor{white} & \textbf{EnrichLog}     & 22.22 & 0.26 & 0.51 & 100.00 & 100.00 & \textbf{100.00} & 99.09 & 100.00 & \underline{99.54} & 99.35 & 100.00 & \underline{99.67} & 96.81 & 100.00 & 98.38 \\
        \rowcolor{gray!20}
       \cellcolor{white} & \textbf{EnrichLog*}    & 20.00 & 0.26 & 0.51 & 100.00 & 100.00 & \textbf{100.00} & 99.87 & 100.00 & \textbf{99.93} & 99.61 & 100.00 & \textbf{99.80} & 98.19 & 100.00 & \underline{99.09} \\
        \bottomrule
    \end{tabular}
    \caption{Performance comparison of different methods across four datasets. 
    \textbf{Base} = pre-trained model. 
    \textbf{Corpus-only} uses only corpus knowledge. 
    \textbf{EnrichLog} and \textbf{EnrichLog*} are one-step and two-step approaches, respectively. 
    Bold values indicate the best F1 for each dataset and model, while \underline{underlined} values indicate the second-best F1 score.}
    \label{tab:overall_accuracy}
\end{table*}

We begin by examining whether the integration of contextual knowledge improves overall performance. To this end, we compare our approach against state-of-the-art techniques for event-based log anomaly detection, including RAG-based methods. 

The results are summarized in Table~\ref{tab:overall_accuracy}.
As shown, pretrained LLMs without external knowledge consistently achieve high recall but suffer from severely low precision, often misclassifying normal log events as anomalies (i.e., high false positives). Increasing model size only partially alleviates this issue. Smaller models tend to label nearly all events as anomalous, whereas larger models (e.g., Mistral-7B) exhibit improved precision but still misclassify many normal logs. These results indicate that reliance only on internal representations introduces systematic bias in anomaly detection.

We observe that augmenting prompts with external knowledge provides explicit guidance and yields substantial improvements in anomaly detection compared to pretrained LLMs. In particular, LogRag and RAGLog, which leverage semantic template information and retrieve reference log entries, enable medium- and large-scale models to achieve notable gains in both precision and F1-score. These methods demonstrate strong performance on the Spirit and Liberty datasets; however, their performance decreases on the more challenging BGL and Thunderbird datasets. We also evaluated LogPrompt in our setting. While it improves over baseline LLM performance, it achieves a similar F1-score compared to LogRag and RAGLog. Additionally, for certain datasets, we observe that RAGLog, LogPrompt, and LogRag fail to detect any true anomalies, resulting in a zero F1-score. Next, we evaluated augmenting the prompt with corpus-level knowledge, where summaries of the log corpus are provided to guide anomaly detection. As seen, the corpus-only technique yields only marginal improvements and suggests that models struggle to directly leverage contextual knowledge for more accurate predictions. 

In contrast, EnrichLog, which incorporates both corpus-level and sample-specific knowledge, consistently outperforms all baseline techniques. Our analysis further reveals that the choice of open-source LLM has a significant impact on performance. Specifically, Mistral-7B achieves the strongest overall results and delivers consistently high F1-scores across all datasets. Interestingly, in certain cases, a smaller model, Gemma-4B, outperforms Mistral-7B, achieving the highest F1-scores on the Thunderbird and Liberty datasets. It is also noteworthy that, despite being comparable in size to Mistral-7B, Llama-8B fails to achieve competitive performance in our setting for Thunderbird. This suggests that model selection can significantly impact anomaly detection performance and should be considered carefully.
We also observe that the Llama-1B model performs poorly with EnrichLog, likely due to its limited capacity to handle longer contexts.

Finally, it is worth noting that EnrichLog*, which uses a two-step approach, achieves performance comparable to or better than EnrichLog. This is because it combines the strengths of high-recall pretrained models with targeted contextual reasoning. Our approach initially identifies potential anomalies using a model biased toward recall and subsequently refines these predictions through contextual knowledge, reducing false positives while maintaining high precision. This design also improves overall efficiency, as log anomaly datasets are typically highly imbalanced with more normal events than anomalies. This allows the system to focus computational resources on the subset of events most likely to be anomalous.

\textbf{Takeaway.} \textit{EnrichLog*, which integrates contextual knowledge, substantially improves log anomaly detection across all datasets and addresses the limitations of pretrained LLMs. Moreover, our approach achieves strong performance with pre-trained open-source LLM models such as Mistral-7B and Gemma-4B, eliminating the need to rely on GPT-like models for event-based anomaly detection.}

\subsection{Impact of Contextual Knowledge Fusion}
\begin{figure}[th]
  \centering
    \includegraphics[width=3.2in]{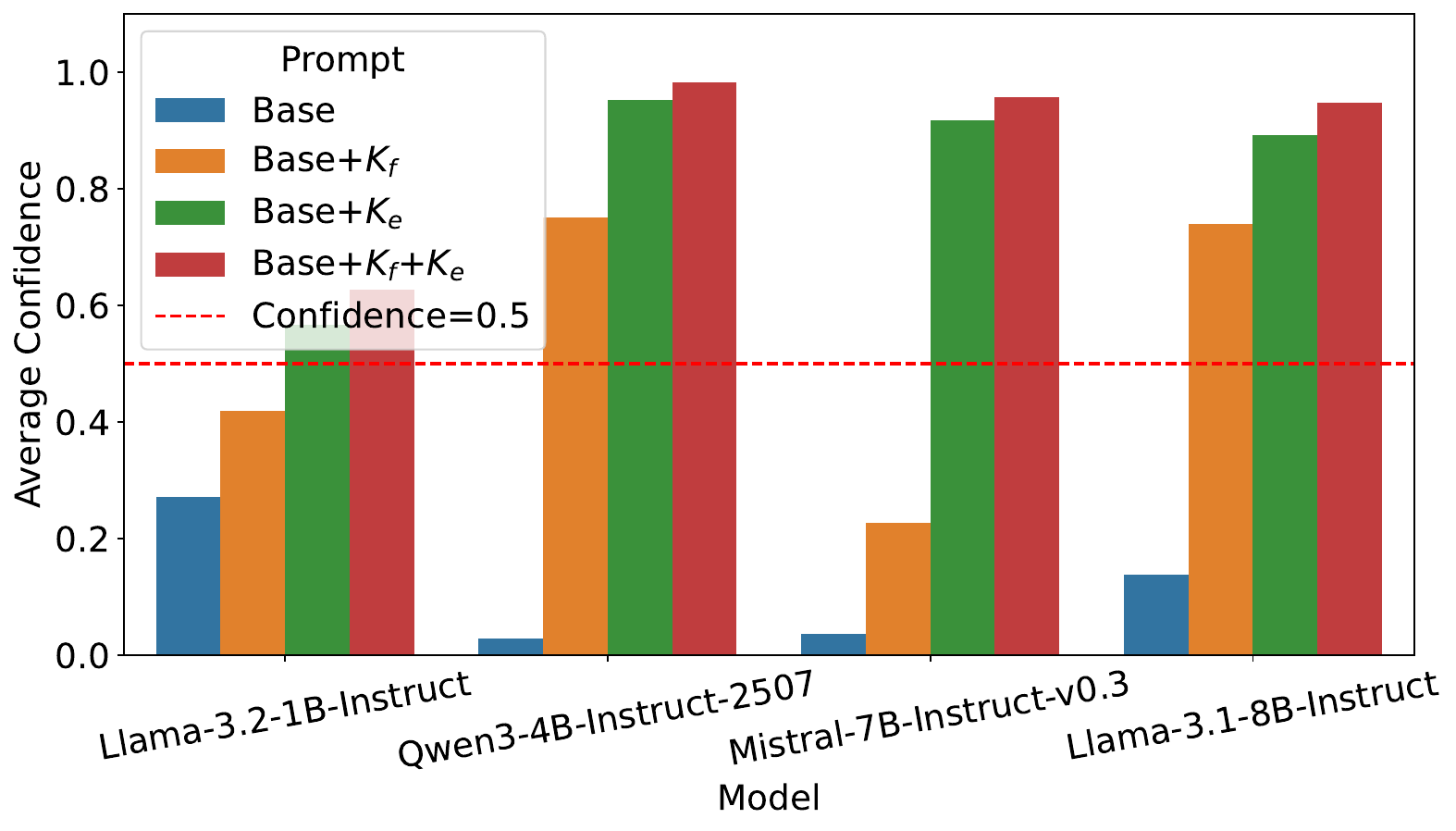} 
   
  \caption{Impact of different contextual knowledge augmentation on model confidence scores.}
  \label{fig:plot_avg}
\end{figure}

We now investigate the effect of augmenting prompts with different forms of contextual knowledge and evaluate the resulting confidence scores. Note that confidence scores represent the probability assigned to a label, which may vary depending on the context provided in the prompt. Specifically, we examine four types of prompts to assess how different sources of external knowledge contribute to classification outcomes: (i) Base only – using only the log entry for anomaly detection, (ii) Base + historical samples ($K_h$) – incorporating historical knowledge from prior logs, (iii) Base + sample-level enriched knowledge ($K_e$) – uses enriched task-specific knowledge, and (iv) Base + both knowledge ($K_h + K_e$) – combining historical and enriched knowledge.

To assess their impact, we focus on instances that were misclassified by the pretrained model and quantitatively measure how each type of external knowledge influences confidence. We set a confidence threshold of 0.5 as a reference point. In other words, a model is considered to have corrected its prediction if the confidence associated with the true label exceeds 0.5. Intuitively, higher confidence values above this threshold indicate that the model successfully shifts from misclassification to correct classification.

Figure~\ref{fig:plot_avg} presents the comparative performance across models. Two key findings emerge. First, the inclusion of contextual knowledge consistently increases both confidence and overall classification accuracy. Second, when comparing individual forms of augmentation, sample-level enrichment ($K_e$) yields greater performance gains than historical knowledge alone ($K_h$). Nonetheless, the combination of both ($K_h + K_e$) achieves the highest overall confidence. We further observe that model size plays a significant role: larger models achieve higher confidence improvements compared to smaller counterparts.

\textbf{Takeaway.} \textit{Augmenting prompts with contextual knowledge substantially improves confidence and classification accuracy on previously misclassified instances. Among the different strategies, sample-level enriched knowledge  provides stronger gains than historical knowledge, while combining both yields the highest overall improvements.}

\subsection{Analysis of Knowledge Enrichment Models}

\begin{table}[t]
    \centering
    \small
    \begin{tabular}{lccc ccc}
        \toprule
        \multirow{2}{*}{Models} & \multicolumn{3}{c}{Qwen3-4B} & \multicolumn{3}{c}{Mistral-7B} \\
        \cmidrule(lr){2-4} \cmidrule(lr){5-7}
         & P & R & F1 & P & R & F1 \\
        \midrule
        \hspace*{-3mm}GPT-5 mini&99.36&98.11&98.73&99.37&99.37&99.37\\
         \hspace*{-3mm}Claude 3.5 Haiku&100.00&99.37&99.68&88.76&99.37&93.77\\
         \hspace*{-3mm}Gemini 2.5 Flash&99.37&99.37&99.37&100.00&99.37&99.68\\
         \hspace*{-3mm}Llama-3.1-8B-Instruct&96.93&99.37&98.14&83.87&98.11&90.43\\
         \hspace*{-3mm}Llama-3.2-1B-Instruct&65.24&95.60&77.55&90.29&99.37&94.61\\
        \bottomrule
    \end{tabular}
    \caption{Comparison of different LLMs for generating sample-specific enriched knowledge on the BGL dataset.}
    \label{tab:enrichment_accuracy}
\end{table}

We investigate the impact of different large language models (LLMs) on generating enriched sample-level knowledge. As described earlier, sample-specific knowledge is first enriched using an LLM, which is subsequently utilized for anomaly detection. The quality of this enrichment depends on the capabilities of the underlying model, and we aim to understand how different models affect the resulting enriched data. To this end, we evaluate five LLMs: three widely recognized commercial models, namely, ChatGPT-5-mini (OpenAI), Claude 3.5-Haiku (Anthropic), and Gemini 2.5-Flush (Google), and two open-source local models, namely, Llama-3.1-8B-Instruct and Llama-3.2-1B-Instruct \cite{grattafiori2024llama}, which represent large-scale and small-scale baseline models, respectively.


Table~\ref{tab:enrichment_accuracy} presents the enrichment results obtained using two different LLM models on the BGL dataset. Overall, sample-level reasoning knowledge generated by commercial models can be effectively utilized. This is likely because large commercial models are trained on extensive and diverse datasets, enabling them to generalize across a wide range of contexts and generate high-quality, sample-specific reasoning knowledge. The exception is the Claude 3.5 + Mistral-7B pairing, which achieves an F1 score of 93\%. 
Furthermore, Llama-3.1-8B, a larger open-source model, also performs well, achieving an F1 score of 98.1\%, demonstrating that effective enrichment can also be performed locally. Interestingly, while smaller enrichment models generally perform worse on Qwen-4B, we observe that the smallest model outperforms Llama-8B when paired with Mistral-7B. These findings suggest that smaller models can be effective for enrichment in specific pairings, although this effectiveness must be evaluated on a case-by-case basis. Nevertheless, our results consistently show that stronger models used for sample-specific enrichment tend to achieve higher overall performance.

\textbf{Takeaway}. \textit{Commercial LLMs and sufficiently large open-source models can effectively generate sample-specific enriched knowledge, while smaller models may perform well in selective pairings. Overall, stronger models consistently yield higher enrichment quality and downstream performance.}

\subsection{Effect of Corpus-Level Knowledge} 
\begin{table}[t]
    \centering
    \small
    \begin{tabular}{lccc ccc}
        \toprule
        \multirow{2}{*}{Corpus Knowledge} & \multicolumn{3}{c}{Qwen3-4B} & \multicolumn{3}{c}{Mistral-7B} \\
        \cmidrule(lr){2-4} \cmidrule(lr){5-7}
         & P & R & F1 & P & R & F1 \\
        \midrule
        None     & 92.86 & 98.11 & 95.41 & 97.53 & 99.37 & 98.44 \\
        Takeaway & 99.36 & 98.11 & 98.73 & 99.37 & 99.37 & 99.37 \\
        Full     & 88.14 & 98.11 & 92.86 & 87.22 & 98.74 & 92.63 \\
        \bottomrule
    \end{tabular}
    \caption{Impact of different forms of corpus-specific knowledge on sample-specific enrichment.}
    \label{tab:knowledge_accuracy}
\end{table}


We now evaluate the effectiveness of corpus-specific knowledge in improving anomaly detection performance. Corpus-specific knowledge provides dataset-specific context during the enrichment process, enabling the generation of sample-specific knowledge that is closely aligned with the characteristics of the dataset. This context helps the enrichment model better identify anomalies in log events and produce relevant enriched knowledge. To investigate this, we focus on three settings for generating sample-specific knowledge: (i) None – no external knowledge is provided for enrichment, (ii) Concise takeaway – only a brief, summarized version of the corpus-specific knowledge is used for enrichment, and (iii) Full – the complete corpus-specific knowledge is provided during enrichment.

Table~\ref{tab:knowledge_accuracy} presents the results across different scenarios. We observe that using full corpus-level knowledge results in the lowest performance. This is because full corpus content contains detailed, technical analyses of log events, which can confuse the open-source models employed in our work. In contrast, omitting external knowledge entirely forces the model to rely on its internal knowledge, yielding moderately better performance in knowledge enrichment. Finally, using concise takeaways for sample-specific knowledge enrichment allows the models to generate focused, scenario-relevant knowledge that is easily processed by the detection LLM. Consequently, concise takeaways provide the most effective form of enrichment for improving detection accuracy.




\textbf{Takeaway}. \textit{Augmenting corpus-specific knowledge for sample-specific enrichment improves overall anomaly detection performance. Notably, concise scenario-specific summaries are more effective than using the full corpus or no external knowledge, likely due to the model sizes employed in this work.}


         
         
         

\subsection{Impact of Embedding Models}



We next evaluate the impact of different embedding models on the RAG process. The choice of embedding model directly affects retrieval quality, which can in turn influence anomaly detection performance. In our previous experiments, we employed a small embedding model (Qwen3-0.6B~\cite{Zhang2025Qwen3EA}) for efficient retrieval. Here, we extend the evaluation to include three additional widely used retrievers: Qwen3-4B~\cite{Zhang2025Qwen3EA}, Contriever~\cite{izacard2021unsupervised}, and multilingual-e5-large~\cite{Wang2024MultilingualET}.

As shown in Table~\ref{tab:embedding_accuracy}, varying the embedding model produces only marginal differences in performance, a trend that holds across other datasets as well. This suggests that anomalous and normal log entries are typically highly dissimilar, allowing even lightweight embedding models to effectively retrieve the most relevant reference logs. These retrieved logs generally share structural patterns or templates with the query log event, making retrieval robust to the choice of embedding model.

\textbf{Takeaway} \textit{Our EnrichLog framework performs effectively even with lightweight embedding models, as anomalous and normal logs are typically highly dissimilar, allowing accurate retrieval of relevant reference logs.}
\begin{table}[t]
    \centering
    \small
    \begin{tabular}{lccc ccc}
        \toprule
        \hspace*{-4mm}\multirow{2}{*}{Embedding Models} & \multicolumn{3}{c}{Qwen3-4B} & \multicolumn{3}{c}{Mistral-7B} \\
        \cmidrule(lr){2-4} \cmidrule(lr){5-7}
         & P & R & F1 & P & R & F1 \\
        \midrule
        \hspace*{-4mm}Qwen3-Embedding-0.6B&99.36&98.11&98.73&99.37&99.37&99.37\\
        \hspace*{-4mm}Qwen3-Embedding-4B&100.00&98.11&99.05&93.45&98.74&96.02\\
        \hspace*{-4mm}Contriever&99.36&98.11&98.73&99.37&99.37&99.37\\
       \hspace*{-4mm}Multilingual-e5-large&98.73&98.11&98.42&98.74&98.74&98.74\\
        \bottomrule
    \end{tabular}
    \caption{Performance on different embedding models.}
    \label{tab:embedding_accuracy}
\end{table}


         
         
         
         

\subsection{Effect of Log Parsers on Enrichment}
\begin{table}[t]
    \centering
    \small
    \begin{tabular}{lccc ccc}
        \toprule
        \multirow{2}{*}{Parser} & \multicolumn{3}{c}{Qwen3-4B} & \multicolumn{3}{c}{Mistral-7B} \\
        \cmidrule(lr){2-4} \cmidrule(lr){5-7}
         & P & R & F1 & P & R & F1 \\
        \midrule
        Drain&99.37&99.37&99.37&99.37&99.37&99.37\\
         Spell&92.44&100.00&96.07&95.18&99.37&97.23\\
         AEL&94.55&98.11&96.30&98.75&99.37&99.06\\
        \bottomrule
    \end{tabular}
    \caption{Comparison of log parsers for sample-specific knowledge enrichment.}
    \label{tab:parser_accuracy}
\end{table}
We next evaluate the impact of log parsers on sample-based enrichment. The log parser is responsible for clustering log entries and generating log templates, which are subsequently used to produce sample-specific enriched knowledge. As such, these extracted templates affect the enrichment knowledge process. To study this, we evaluate three widely used parsers: Spell\cite{Du2016SpellSP}, Drain~\cite{He2017DrainAO}, and AEL\cite{Jiang2008AbstractingEL}. Due to computational and cost constraints, we focus on enriching only frequently occurring templates and  consider only those that appear more than 10 times in the dataset. After filtering out rare templates, the remaining counts are 659 for Spell, 344 for Drain, and 320 for AEL. These log templates are then enriched using GPT-5 mini and stored in a knowledge base for our analysis.

Table~\ref{tab:parser_accuracy} compares the performance of these parsers. As shown, Drain achieves the best overall performance. Interestingly, the number of extracted templates, whether higher or lower, does not appear to strongly correlate with downstream anomaly detection performance. This suggests that the quality of the log template clustering technique, rather than the quantity of log templates, plays a key role in the effectiveness of enrichment.



\textbf{Takeaway}  \textit{The choice of parser influences performance, with Drain achieving the best performance.}


         
         
         
         
\subsection{Impact of Ambiguous Log Templates} 

\begin{table}[t]
    \centering
    \small
    \begin{tabular}{lcccc cccc}
        \toprule
        \multirow{2}{*}{} & \multicolumn{3}{c}{Qwen3-4B} & \multicolumn{3}{c}{Mistral-7B} \\
        \cmidrule(lr){2-4} \cmidrule(lr){5-7}
         & P & R & F1 & P & R & F1 \\
        \midrule
        EnrichLog*&92.69&88.24&90.41&92.81&91.43&92.12\\
        RAGLog&77.60&99.96&87.37&65.95&100.00&79.48\\
        LogPrompt&15.13&3.06&5.10&50.12&99.75&66.72\\


        \bottomrule
    \end{tabular}
    \caption{Performance on ambiguous log templates.}
    \label{tab:special_accuracy}
\end{table}
We next evaluate the impact of ambiguous templates, where both anomalous and normal log entries may be mapped to the same log template. To systematically study these cases, we use a subset of the BGL test dataset containing such ambiguities. Specifically, this subset includes 2,415 anomalous logs and 2,766 normal logs that are assigned to templates containing entries from both classes.

Table~\ref{tab:special_accuracy} presents our results. Despite the inherent ambiguity, our framework achieves 92\% F1-score and outperforms baseline techniques, demonstrating its ability to accurately distinguish anomalies even when multiple log classes are mapped to the same template. These results highlight the robustness of sample-specific enrichment and leveraging raw logs in handling ambiguous cases.



\textbf{Takeaway}. \textit{EnrichLog* enables robust anomaly detection even in cases where log templates are ambiguous.}

\subsection{Efficiency Analysis of Two-Step Approach}
\begin{table}[t]
    \centering
    \small
    \begin{tabular}{l c cc cc}
        \toprule
        \multirow{2}{*}{Models} & LLM (s) & \multicolumn{2}{c}{One-step (s)} & \multicolumn{2}{c}{Two-step (s)} \\
        \cmidrule(lr){2-2} \cmidrule(lr){3-4} \cmidrule(lr){5-6}
         & Gen & Gen & Total & SS-Gen & Total \\
        \midrule
        Qwen3-4B (int4)   & 0.0630 & 0.0713 & 0.1178 & 0.1026 (90\%) & 0.1665 \\
        Qwen3-4B (fp16)   & 0.0659 & 0.2072 & 0.2532 & 0.2275 (91\%) & 0.3047 \\
        Mistral-7B (int4) & 0.0892 & 0.1429 & 0.1889 & 0.0700 (35\%) & 0.1619 \\
        Mistral-7B (fp16) & 0.1393 & 0.4249 & 0.4725 & 0.0817 (15\%) & 0.2253 \\
        \bottomrule
    \end{tabular}
    \caption{Efficiency comparison of OS and TS settings across different models. 
    ``Gen": generation time, ``SS-Gen": average generation time in the second step, and percentages indicate the proportion of cases executed in the second step.}
    \label{tab:speed_time}
\end{table}

         

\begin{table}[t]
    \centering
    \small
    \begin{tabular}{lccc ccc}
        \toprule
        \multirow{2}{*}{Models} & \multicolumn{3}{c}{One-step} & \multicolumn{3}{c}{Two-step} \\
        \cmidrule(lr){2-4} \cmidrule(lr){5-7}
         & P & R & F1 & P & R & F1 \\
        \midrule
        Qwen3-4B (int4)&95.1&99.3&97.2&94.6&99.3&96.9\\
         Qwen3-4B (fp16)&99.3&98.1&98.7&99.3&98.1&98.7\\
         Mistral-7B (int4)&99.3&96.8&98.0&98.7&99.3&99.0\\
         Mistral-7B (fp16)&99.3&98.1&98.7&99.3&99.3&99.3\\
        \bottomrule
    \end{tabular}
    \caption{Impact of quantization on detection performance.}
    \label{tab:speed_accuracy}
\end{table}

We next compare the overall efficiency of the two-step (EnrichLog*) approach. To do so, we evaluate both the detection inference time --- the time taken by the LLM to produce an output given a prompt---and the \textit{total} time, which includes the RAG pipeline for retrieving relevant documents. Specifically, we use \textit{Gen} to denote the time required for token generation in the first step, and \textit{SS-Gen} to denote the generation time in the second step of the two-step algorithm. Some log entries may skip the second step. In these cases, SS-Gen is recorded as zero, and we report the percentage of log entries that execute the second step. A lower percentage indicates that more entries skip the second step, leading to greater efficiency.

Table~\ref{tab:speed_time} summarizes the results for the one- and the two-step approach across different quantized models. We also report the performance of LLM, which EnrichLog* uses in the first step of its pipeline. We observe that pretrained LLM inference requires 65 ms on Qwen3-4B and 140 ms on Mistral-7B (using FP16 precision). Using INT4 quantization, the inference time reduces to 63 ms for Qwen3-4B and 50 ms for Mistral-7B. Since this task is fundamentally a Time-To-First-Token (TTFT) problem, quantization does not always yield proportional speedups.

We further observe that overall time increases significantly when incorporating the RAG pipeline, with generation time rising by 214\% for Qwen3-4B and 205\% for Mistral-7B.
Compared with the two-step approach, the one-stage (OS) method exhibits higher end-to-end inference time for Mistral. This is because the two-step (TS) approach incorporates a fast and slow pass. In the first step, it uses only the basic pretrained LLM, which provides fast inference but lower precision. The second step is equivalent to the OS approach, employing the RAG pipeline for anomaly detection. However, for Qwen3-4B, approximately 91\% of new logs enter the slow pass, resulting in an overall runtime that exceeds that of OS. This is because Qwen3-4B is a smaller model with limited capacity to filter normal logs during the fast pass. In contrast, Mistral-7B, being a larger model, admits only 15\% of new logs to the slow pass, demonstrating greater efficiency in filtering and reducing overall runtime.

We further evaluate the impact of quantization on detection performance. As shown in Table~\ref{tab:speed_accuracy}, quantization reduces the F1-score by approximately 0.3 to 2.0 points.
At the same time, it significantly accelerates inference, achieving speedups of 28\% to 60\% depending on the model and precision used. These results highlight a clear trade-off between accuracy and efficiency, and show that quantization can be an effective strategy for improving inference speed in resource-constrained settings, although with a slight loss in detection performance.


To assess practical performance, we evaluate the most promising configurations on the full BGL test dataset, which contains over 900,000 log lines. The results indicate that Qwen (FP4) achieves Precision = 89.16, Recall = 99.67, F1 = 94.13, whereas Mistral (FP16) achieves Precision = 98.67, Recall = 99.65, F1 = 99.16, demonstrating that larger models maintain high accuracy even on large-scale datasets while offering efficient inference when appropriately configured.

\textbf{Takeaway:} \textit{The two-step (EnrichLog*) approach improves overall efficiency by using a fast initial pass to filter normal logs, reducing the number of samples processed by the slower, retrieval-augmented second stage. Further, quantization provides significant inference speedups with only minor reductions in detection accuracy.}

\section{Discussion} 

Traditional template-based log anomaly detection approaches rely on parsed template IDs and are susceptible to ambiguous templates.
EnrichLog does not require exact template matches and improves the classification accuracy of ambiguous logs. Empirical results show that incorporating both sample- and corpus-specific knowledge enables the model to correctly classify most log entries.  

Despite its advantages, several challenges remain. 
First, methods still struggle in dynamic environments, particularly when log formats or content evolve over time. EnrichLog currently enriches new logs using the nearest available knowledge; however, completely unseen logs may retrieve irrelevant knowledge, leading to misclassification. Introducing a threshold may mitigate this issue, but as demonstrated by the baseline and corpus-only results, pretrained models often fail to correctly classify logs without external knowledge. 
Thus, there is still scope for incorporating fixed in-domain knowledge to enhance baseline accuracy.


Second, detecting anomalies across systems remains a key challenge. While the knowledge base can store knowledge from multiple datasets, retrieval remains problematic. EnrichLog currently retrieves knowledge using only log events, without incorporating additional system-specific information. Because similar logs may carry different severity levels across systems, they may correspond to different labels. Without system-specific information, retrieval cannot be guaranteed to be reliable. Developing retrieval mechanisms that incorporate system-specific signals therefore represents a good direction for future work.

\section{Conclusion}
This work explores the use of contextual knowledge for event-based anomaly detection in system logs with off-the-shelf pretrained LLMs. We employ RAG techniques to adapt generative models to domain-specific log detection without additional training. To improve both detection performance and efficiency, we introduce EnrichLog, a framework that leverages multiple sources of knowledge, including historical logs, sample-specific enrichment, and corpus-level knowledge, to enhance anomaly detection with pretrained LLMs. We further design a two-step (TS) inference pipeline that accelerates detection while preserving high accuracy.  Our results demonstrate that EnrichLog consistently achieves high F1 scores across all evaluated datasets and that incorporating contextual knowledge substantially improves detection performance. Additionally, the two-step (TS) inference pipeline can deliver significant speedups over the one-step (OS) approach by reducing inference time on both fp16 models and int4 models. Our findings also highlight the robustness of EnrichLog in handling ambiguous log entries, making it suitable for practical deployments. 


\clearpage
\bibliography{bibfile}
\bibliographystyle{acm}

\end{document}